\documentclass[journal]{IEEEtran}
\usepackage{amsmath,amsfonts}
\usepackage{algorithmic}
\usepackage{algorithm}
\usepackage{array}
\usepackage[caption=false,font=footnotesize]{subfig}
\usepackage{textcomp}
\usepackage{stfloats}
\usepackage{placeins}
\usepackage{url}
\usepackage{verbatim}
\usepackage{graphicx}
\usepackage{cite}
\usepackage{booktabs}
\usepackage{multirow}
\usepackage{amssymb}
\usepackage{hyperref}
\newlength{\featurecellwidth}

\begin{document}

\title{Revealing Inter-Channel Modality Discriminability: Channel Selective Interaction Fusion Network for Optical-SAR Object Detection}

\author{Haojie~Li,
        Yuanfei~Huang,~\IEEEmembership{Member,~IEEE,}
        Lingfei~Song,
        Jia~Li,
        Lizhi~Wang,~\IEEEmembership{Member,~IEEE}
\thanks{The authors are with the School of Artificial Intelligence, Beijing Normal University; Beijing Key Laboratory of Artificial Intelligence for Education; Engineering Research Center of Intelligent Technology and Educational Application, Ministry of Education, Beijing 100875, China. Email: 202521081034@mail.bnu.edu.cn, \{yfhuang, lingfei, jiali, lizhiwang\}@bnu.edu.cn. (\textit{Corresponding author: Yuanfei Huang})}}

\markboth{Journal of \LaTeX\ Class Files,~Vol.~14, No.~8, August~2021}%
{Shell \MakeLowercase{\textit{et al.}}: A Sample Article Using IEEEtran.cls for IEEE Journals}


\maketitle

\begin{abstract}
Synthetic aperture radar (SAR) and optical imagery provide complementary structural and appearance information for remote sensing object detection. However, substantial cross-modal discrepancies and indiscriminate channel interaction inevitably introduce redundant or conflicting features. 
Since different feature channels encode heterogeneous semantic information and contribute unequally to cross-modal fusion, selectively highlighting task-relevant channels while suppressing redundant ones is essential.
To address this issue, we propose the Channel Selective Interaction Fusion Network (CSIF-Net), which performs selective channel-level cross-modal interaction. CSIF-Net first employs a frequency decomposition module to construct frequency-aware SAR inputs, facilitating the separation of structural information from noise-related responses. It then models cross-modal channel relationships through a cross-attention channel selection module and dynamically partitions the features into shared and modality-specific components. Finally, a dual-branch fusion module performs cross-modal interaction and channel realignment for shared features while independently enhancing SAR-specific features, thereby exploiting complementary information and suppressing modality interference. Extensive experiments on the OGSOD 1.0 and M4-SAR datasets demonstrate that CSIF-Net achieves state-of-the-art performance in both horizontal and oriented object detection, surpassing the existing methods by 4.01\% and 3.18\% gains, respectively.
\end{abstract}


\section{Introduction}
\IEEEPARstart{S}{ynthetic} aperture radar (SAR) imagery is sensitive to the electromagnetic properties of targets, enabling reliable imaging under diverse illumination and weather conditions but providing limited visual appearance information. In contrast, optical imagery contains clearer visual representations, but its quality is more susceptible to illumination and weather conditions \cite{1_Liu2025IVIFsurvey,18_Zhang2026MultimodalFO}. Therefore, SAR and optical imagery are highly complementary, and provide more comprehensive target representations \cite{3_VSFF20243,5_Liu2024OpticalSARReview,7_DomainAdaptive2024,31_LI2022Review,32_VivoneRSFReview}. Nevertheless, effective optical-SAR fusion remains challenging because of the substantial discrepancies between the two modalities, including differences in imaging mechanisms and visual appearance \cite{3_VSFF20243,5_Liu2024OpticalSARReview,8_DAOOD2025,20_Li2021DTCDN,9_LI2023Pgfusion,11_Gao2025DEN}. These discrepancies may hinder the extraction of consistent target representations and introduce irrelevant or conflicting information during feature fusion.

Existing studies mainly explore two directions of modality translation and cross-modal fusion. 
For modality translation, image translation and generative approaches are widely adopted \cite{8_DAOOD2025,14_Li2021AttentionFGAN,20_Li2021DTCDN,45_Zhang2023TransformerBasedConditionalGAN}. These methods employ generative models to transform heterogeneous data into a unified modality space, thereby facilitating knowledge transfer and reducing feature discrepancies caused by different imaging mechanisms. However, the generated representations are often optimized for visual consistency rather than downstream detection objectives, limiting their ability to fully exploit complementary cross-modal information.
For cross-modal fusion, researchers have developed various strategies to enhance feature interaction and alignment across modalities, developing domain adaptation and cross-modal alignment methods \cite{4_LiuRepL2025,7_DomainAdaptive2024,9_LI2023Pgfusion,16_Li2025MulFS-CAP,17_Ren2025DCIFNet}. Dual-stream backbone architectures \cite{10_REN2022DDHRNet,15_Ma2024MultilevelFusion} further enable hierarchical feature extraction and interaction between SAR and optical modalities. Besides, attention-based methods \cite{6_ICAFusion2024,11_Gao2025DEN,13_Tang2023YDTR,33_JiangCGPIGF,40_Li2024MultispectralPedestrianDetection,41_Li2023ConfidenceAwareFusion,42_Chen2024HitFusion,44_Wang2023InfraredVisibleImageFusion} adaptively model cross-modal feature dependencies to enhance complementary information while suppressing conflicting responses. More recently, Mamba-based architectures \cite{12_FusionMamba2025} have shown promising capability in capturing long-range cross-modal dependencies, while gated fusion mechanisms \cite{3_VSFF20243,17_Ren2025DCIFNet,33_JiangCGPIGF} dynamically regulate information flow during feature aggregation.

\begin{figure*}[!t]
\centering
\subfloat[Averaged accumulated gradients during the training process]{\includegraphics[height=2.3in]{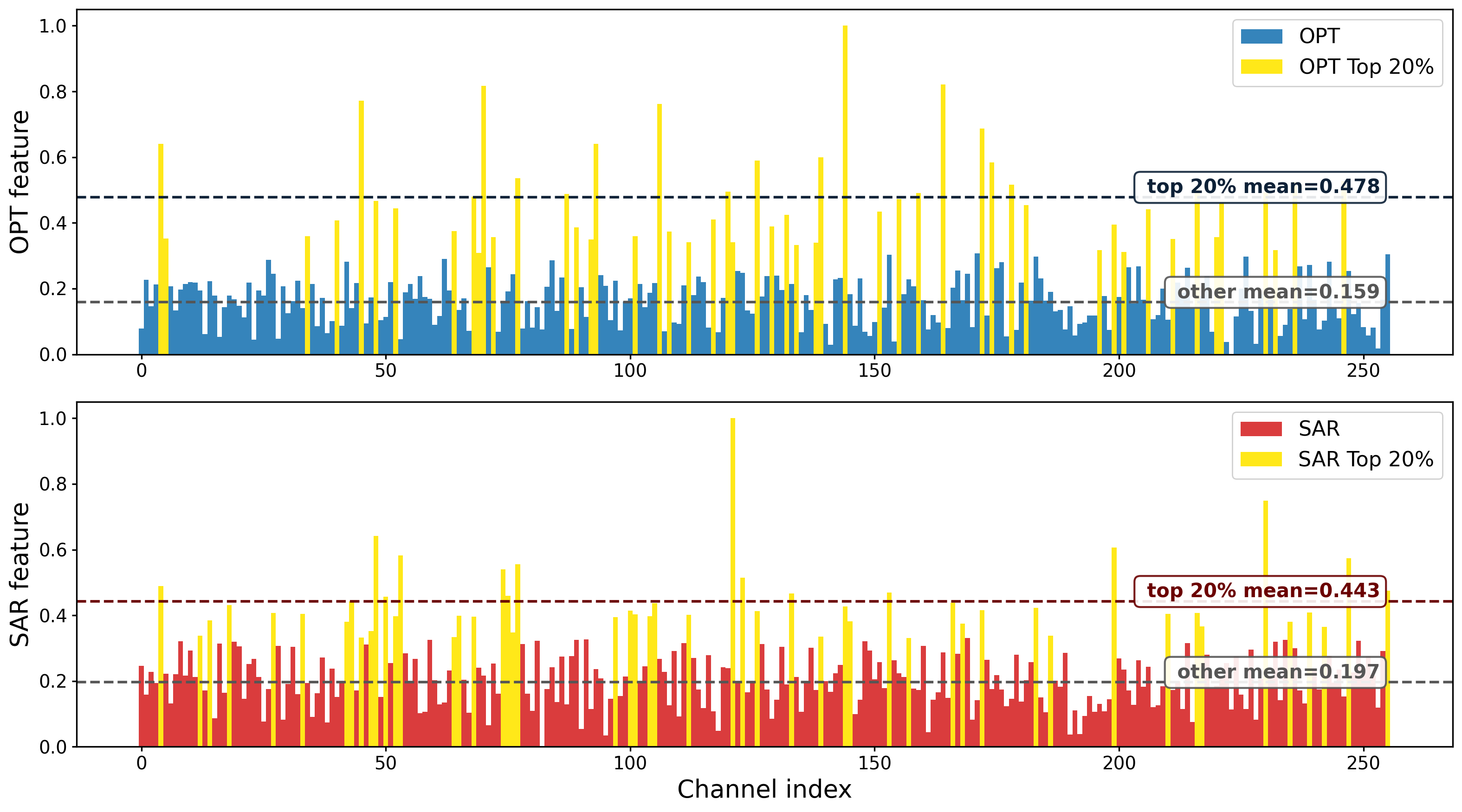}
}
\hspace{0.2in}
\subfloat[Cross-modal channel correlation matrix]{\includegraphics[height=2.3in]{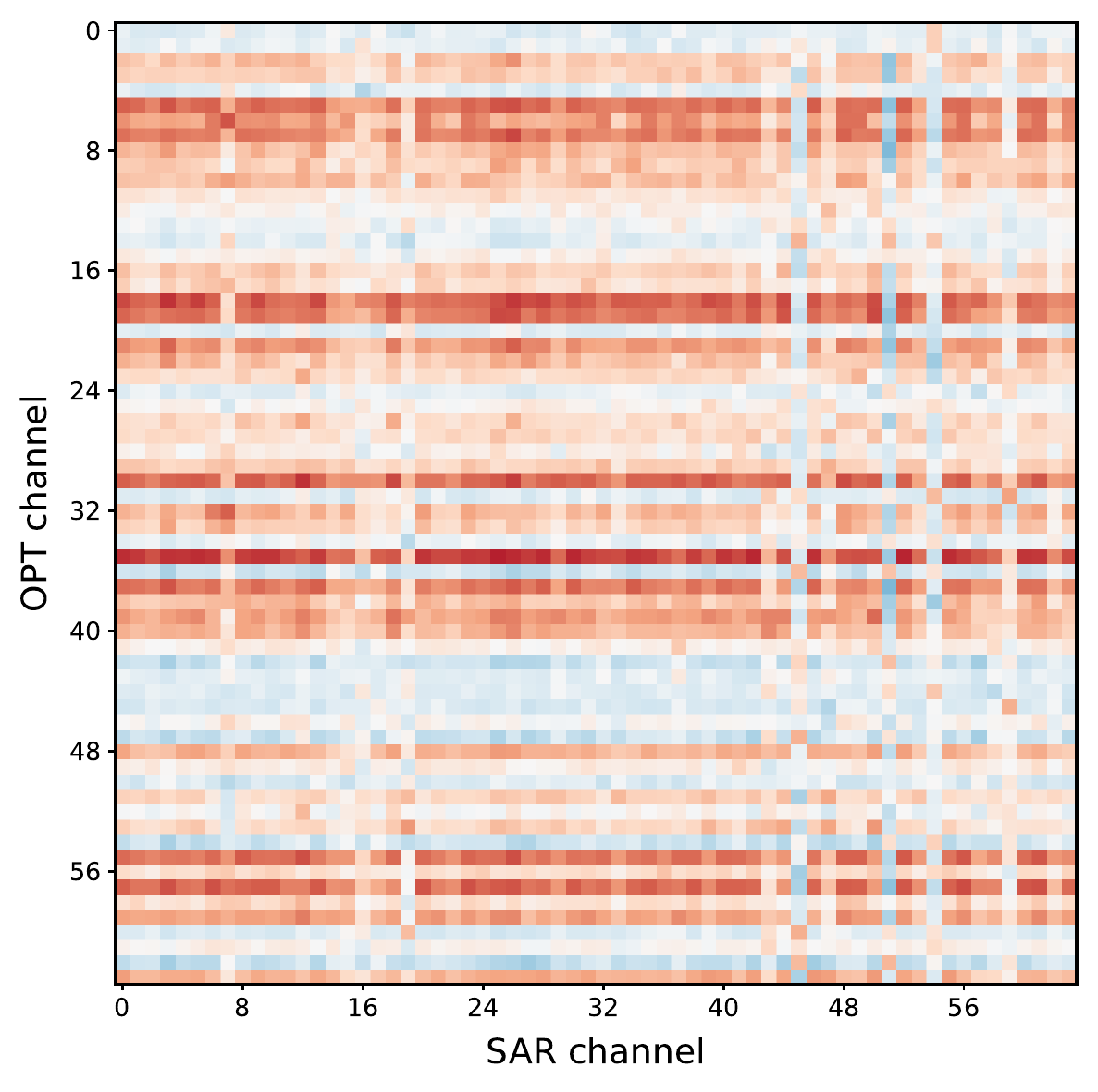}
} 
\caption{We employ a simple 1$\times$1 convolutional network for fusion and object detection, and track the accumulated gradients of each channel during training. The top-64 channels with the largest accumulated gradients are selected from the OPT and SAR modalities respectively to compute their channel correlation matrix. As can be observed, the accumulated gradients vary substantially across channels. Moreover, the cross-modal channel correlation matrix exhibits a pronounced banded pattern, with significant differences in correlation across channels.}
\label{fig_sim}
\end{figure*}

Although existing cross-modal feature alignment and fusion methods have partially alleviated modality interference, they typically model feature interactions mainly in the spatial domain and treat channel-wise features with either uniform importance or unimodal-driven attention. They lack an explicit cross-modal view of which channels carry shared, complementary, or modality-specific information \cite{37_Yang2026DWSFNet,43_Shi2025Frefusion}. However, as illustrated in Fig.~\ref{fig_sim}, {\bf\em different feature channels often encode heterogeneous semantic information, exhibit varying cross-modal attention responses, and contribute unequally to downstream detection}. Indiscriminately aggregating all channels may introduce redundant representations and modality-specific interference, thereby weakening the effectiveness of cross-modal interaction~\cite{19_cao2023CSSA}. Therefore, selectively filtering and dynamically interacting with features at the channel level is crucial for emphasizing task-relevant information and fully exploiting the complementary characteristics of SAR and optical modalities. 

To address the above issues and serve the SAR and optical cross-modal object detection task, we propose a feature-level fusion framework, called Channel Selective Interaction Fusion Network (CSIF-Net). First, a Fast Fourier Transform (FFT)-based Frequency Decomposition Module (FDM) separates the SAR input into multiple frequency components, facilitating the subsequent learning of structural and noise-related patterns. Next, the Cross-Attention Channel Selection Module (CACS) performs the sole channel embedding operation and constructs a cross-modal attention matrix to partition the channels into shared and specific sets. A straight-through estimator (STE) enables this discrete selection process to be optimized end-to-end. Finally, the Dual-Branch Fusion Module (DBF) applies differentiated processing to the indexed components by sharing features that undergo cross-modal interaction and channel alignment, while the SAR specific features are enhanced by self-attention before reconstruction. By jointly leveraging CACS and DBF, CSIF-Net identifies cross-modal channel relationships and performs differentiated feature interaction, thereby enhancing complementary information while suppressing channel redundancy and modality-specific interference.
The main contributions of this work are summarized as follows,
\begin{itemize}
    \item We analyze the channel discrepancy in cross-modal fusion, revealing that different feature channels exhibit highly imbalanced cross-modal attention responses and contribute unequally to downstream detection, emphasizing the necessity of channel-level selective interaction. 
    \item We propose a cross-modal selective fusion framework for optical-SAR object detection, which achieves channel-level cross-modal feature disentanglement and dynamic interaction through a cross-modal attention matrix.
    \item We design distinct strategies for shared and specific features, where shared features are fused via cross-attention with channel realignment, while the CACS embeddings of SAR specific features undergo self-attention and reconstruction.
\end{itemize}

\section{Related Work}
\subsection{SAR and Optical Cross-Modal Fusion Strategies}
Recent surveys have summarized multimodal remote sensing fusion from the perspectives of fusion levels, protocols, datasets, and downstream applications \cite{31_LI2022Review,32_VivoneRSFReview}. Owing to their complementarity, optical-SAR fusion has been widely studied in land cover classification, semantic segmentation, building extraction, and heterogeneous change detection. Existing studies have explored different strategies to bridge the modality gap between SAR and optical images. Liu et al. \cite{5_Liu2024OpticalSARReview} reviewed deep feature fusion methods for optical and SAR semantic segmentation, emphasizing the importance of multimodal feature representation and cross-modal alignment. For land cover and land use classification, Ren et al. \cite{10_REN2022DDHRNet} proposed DDHRNet, which adopts a dual-stream high-resolution architecture to fuse GF-2 optical and GF-3 SAR features and improves classification performance under degraded optical conditions. MCANet \cite{22_LIMCANet} further introduces pseudo-Siamese feature extraction, multimodal cross-attention, and multi-level feature fusion for joint optical-SAR land use classification. Optical-SAR fusion has also been investigated in building extraction and change detection. Li et al. \cite{9_LI2023Pgfusion} proposed a progressive fusion learning framework for building extraction, where shared modal-invariant features are extracted as an information bridge for multistage optical-SAR fusion. For heterogeneous change detection, Jiang et al. \cite{21_JiangChangeDetection} introduced deep homogeneous feature fusion to transform optical and SAR images into a comparable feature space while preserving semantic content. Liu et al. \cite{4_LiuRepL2025} proposed commonality feature representation learning to extract latent comparable features from multimodal bi-temporal images for unsupervised change detection.

These methods demonstrate the effectiveness of optical-SAR fusion, but they mainly target scene-level dense prediction tasks such as segmentation, classification, and change detection. Object detection additionally demands precise localization, instance-level discrimination, and robust multi-scale representation, so detection-oriented optical-SAR fusion that suppresses modality interference and exploits complementarity remains an open problem.
\subsection{Cross-Modal Image Fusion for Object Detection}
Recent multimodal remote sensing detectors further explore correlation-guided progressive interaction, gated fusion, diffusion deformation, and knowledge distillation to improve optical-SAR feature complementarity and detection robustness \cite{33_JiangCGPIGF,34_YueDMKD}. Cross-modal object detection aims to improve detection robustness by exploiting complementary information from heterogeneous modalities. In optical-SAR remote sensing, related studies are still relatively limited and mainly focus on cross-domain transfer, few-shot learning, and benchmark construction. Zhou et al. \cite{7_DomainAdaptive2024} proposed a domain-adaptive few-shot SAR ship detection method that transfers knowledge from optical images to SAR images through feature distance measurement and dual-stream feature alignment. Huang et al. \cite{8_DAOOD2025} investigated domain adaptive oriented object detection from optical to SAR images, where pixel-level and instance-level information transfer are jointly used to bridge the modality gap. Yang et al. \cite{23_YangUDAD} further introduced cross-domain feature interaction and data contribution balance for unsupervised SAR ship detection. Recently, Wang et al. \cite{24_WangM4SAR} constructed M4-SAR, a large-scale optical-SAR fusion object detection benchmark, and proposed a Mamba-based optical and SAR fusion algorithm which demonstrated good performance. Since optical-SAR object detection remains underexplored, infrared-visible and multispectral object detection methods provide useful references for general cross-modal detection \cite{38_Zhou2025COFNet,39_Wang2026RGBTObjectDetection,46_Wan2024MFFNet}. Fang et al. \cite{25_FangCFT} proposed a Cross-Modality Fusion Transformer to model intra- and inter-modality interactions between RGB and thermal features. Shen et al. \cite{6_ICAFusion2024} proposed ICAFusion, which uses iterative dual cross-attention transformers to capture complementary information across modalities. Helvig et al. \cite{27_HelvigCAFF} introduced CAFF-DINO by integrating cross-attention feature fusion into a Transformer-based detector.

Nevertheless, these detectors rely mainly on spatial interaction, global attention, or domain adaptation and seldom model channel-level cross-modal relationships, motivating more selective feature interaction.
\subsection{Channel-Level Feature Differential Processing}
Channel-level feature modeling has been widely used to improve representational ability. Hu et al. \cite{28_HuSENet} proposed the squeeze-and-excitation block, which models channel interdependencies to adaptively recalibrate channel responses; Li et al. \cite{35_LiSKNet} introduced Selective Kernel Networks to adaptively select receptive fields; Woo et al. \cite{29_WooCBAM} proposed CBAM, which sequentially infers channel and spatial attention; and Wang et al. \cite{30_WangECANet} proposed ECA-Net, which uses efficient local cross-channel interaction without dimensionality reduction. These methods show that channels contribute unequally and that channel-wise recalibration can improve downstream performance. In multimodal fusion tasks, channel-level operations have also been explored to enhance complementary information and reduce computational cost. Cao et al. \cite{19_cao2023CSSA} proposed a lightweight channel switching and spatial attention module for multimodal object detection, which efficiently exchanges information between infrared and visible modalities while emphasizing informative spatial regions. Yuan et al. \cite{26_YuanC2Former} designed C$^{2}$Former for RGB-infrared object detection, where intermodality cross-attention is used to learn calibrated and complementary features across modalities. These methods indicate that channel-level or attention-based interaction is effective for improving cross-modal detection performance.

However, existing channel attention methods estimate importance from a single modality or globally recalibrate features without distinguishing shared from modality-specific channels in cross-modal fusion. In optical-SAR fusion this distinction is especially important, since SAR features contain both useful scattering structures and noise-related responses. Modeling channel-level cross-modal relationships and differentially processing shared and modality-specific features is therefore necessary for more selective and robust fusion.
\section{Proposed Method}
The objective of cross-modal fusion is to exploit complementary information across modalities while suppressing redundant features. However, due to the substantial differences in imaging mechanisms, directly fusing all feature channels often introduces low-contribution or conflicting information that degrades the fused representation. To address this, we propose a cross-modal fusion framework based on frequency-aware prior enhancement and shared-specific feature disentanglement.

\subsection{Overall Framework}
\begin{figure*}[!t]
\centering
\includegraphics[width=0.85\textwidth]{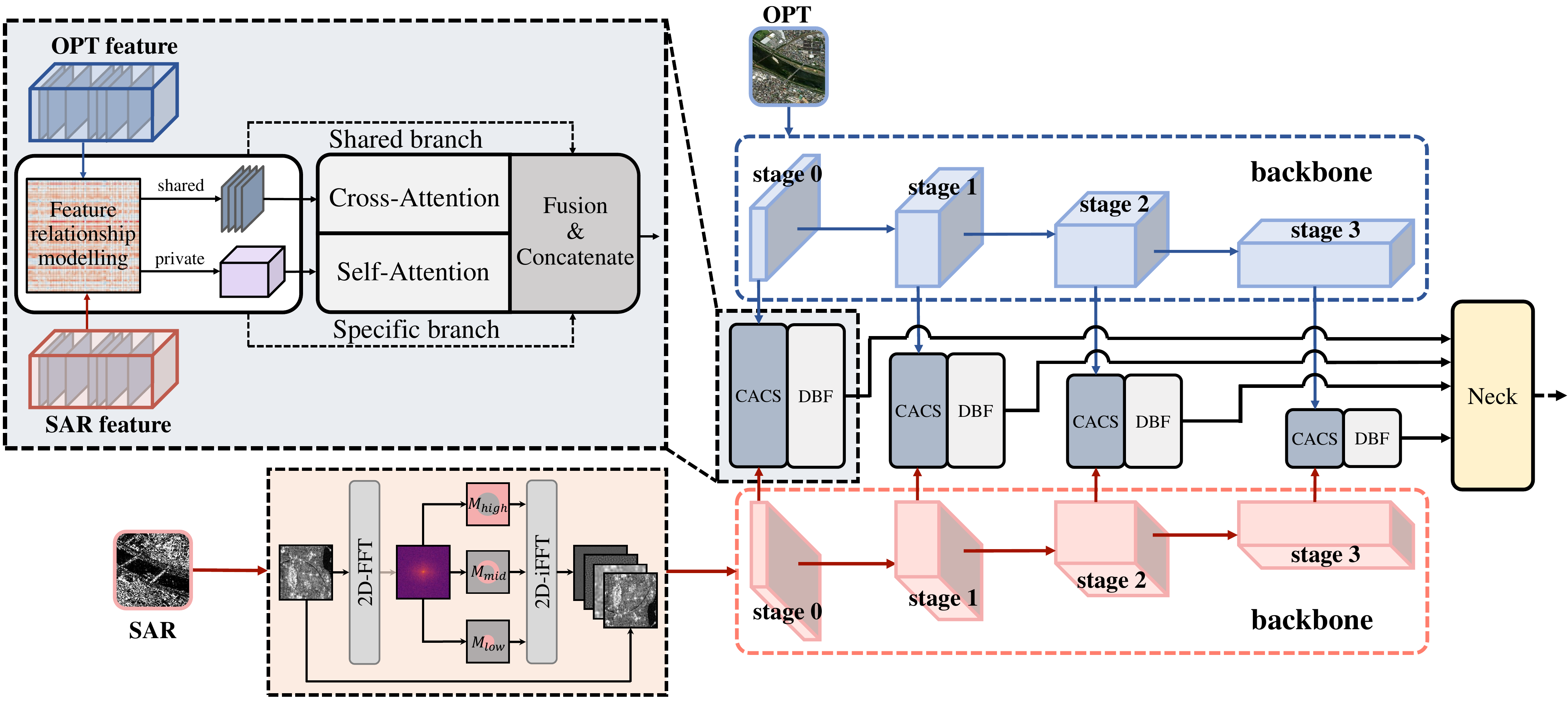} 
\caption{Overall architecture of the proposed CSIF-Net. FDM decomposes the SAR input into multiple frequency components to provide frequency-aware priors for feature extraction. CACS then models channel-wise correlations between the optical and SAR features and partitions them into shared and specific components. Finally, DBF enhances the shared components through cross-modal interaction while preserving modality-specific information in the specific components before multi-scale feature aggregation.}
\label{fig:overall_framework}
\end{figure*}

As illustrated in Fig.~\ref{fig:overall_framework}, the proposed framework consists of three major components: a Frequency Decomposition Module (FDM), a Cross-Attention Channel Selection Module (CACS), and a Dual-Branch Fusion Module (DBF).

Given the optical features $F_{opt}$ and SAR features $F_{sar}$ extracted by the backbone, the overall fusion process can be formulated as

\begin{equation}
F_{out}
=\mathcal{F}_{DBF}
\bigl(F_{opt},F_{sar};\mathcal{F}_{CACS}(F_{opt},F_{sar})\bigr),
\end{equation}
where $\mathcal{F}_{CACS}(\cdot)$ returns the channel embeddings and selection indices $\mathcal I$, while $\mathcal{F}_{DBF}(\cdot)$ retrieves the indexed components and performs dual-branch fusion.

The FDM is placed before the backbone to enrich the frequency representation of SAR inputs, while cross-modal fusion is performed between the backbone and the neck, so that shared information is deeply fused and modality-specific information is preserved before multi-scale aggregation.
The core idea is to explicitly distinguish the roles of channels in cross-modal fusion: some channels encode shared semantics across modalities, while others carry modality-specific observations. Accordingly, CSIF-Net partitions channels into shared and specific sets via a cross-modal attention matrix and processes them differently: shared channels undergo cross-modal interaction and channel alignment, while specific channels are enhanced or retained within their own modality. In the following, we detail the FDM, CACS, and DBF in sequence.
\subsection{Frequency Decomposition Module}
Due to the unique imaging mechanism of SAR, SAR images contain abundant speckle noise, with noise-related responses and structural information highly entangled in the spatial domain. To this end, the Frequency Decomposition Module (FDM) decomposes the SAR input in the frequency domain. A two-dimensional FFT is first applied to the SAR image $X$, and the spectrum is partitioned into low-, middle-, and high-frequency bands using two radial thresholds $r_1<r_2$ and the corresponding binary masks $M_{low}$, $M_{mid}$, and $M_{high}$. Each band is then transformed back to the spatial domain via the inverse FFT, and the resulting components are concatenated with the original image along the channel dimension:
\begin{equation}
    X_{out}=\mathrm{Concat}\left(X_{low},\,X_{mid},\,X_{high},\,X\right),
\end{equation}
where $X_i=\mathcal{F}^{-1}(\mathcal{F}(X)\odot M_i)$ for $i\in\{low,mid,high\}$. The low-, middle-, and high-frequency components preserve global structure, object contours and texture, and fine details together with noise, respectively. By presenting these components as separate inputs, FDM enables the backbone to learn their relative contributions and distinguish useful structures from noise without an explicit denoising operation.
\subsection{Cross-Attention Channel Selection Module}
The relationships between channels from different modalities are inherently heterogeneous. While some channels exhibit clear counterparts across modalities, others primarily describe modality-dependent observations. Therefore, establishing channel-level correspondences before cross-modal interaction is essential for identifying the roles of different channels in the joint representation space. To this end, we propose a Cross-Attention Channel Selection Module (CACS), which constructs a cross-modal attention matrix between optical and SAR features and partitions channels according to their attention responses.

Each channel is first encoded into a compact semantic descriptor. This is the only embedding operation in the proposed framework. Considering the varying spatial resolutions across network stages, high-resolution features are compressed using depthwise convolution before linear projection, whereas low-resolution features are directly projected. The channel embeddings are
\begin{equation}
E_m=\Phi(F_m)\in\mathbb{R}^{C_m\times d},
\quad m\in\{opt,sar\}.
\end{equation}
where $\Phi(\cdot)$ denotes the channel embedding function and $d$ is the embedding dimension.
The embeddings are then used to construct a cross-modal attention matrix:
\begin{equation}
A=
\frac{Q_{opt}K_{sar}^{\top}}{\sqrt d},
\quad A\in\mathbb{R}^{C_{opt}\times C_{sar}},
\end{equation}
where $Q_{opt}$ and $K_{sar}$ are linear projections of $E_{opt}$ and $E_{sar}$, respectively, and the Softmax operation is performed over the SAR-channel dimension. Each element of $A$ represents the cross-modal attention weight of an optical-SAR channel pair. For optical-channel scoring, the attention vectors are obtained row-wise from $A$; for SAR-channel scoring, they are obtained column-wise from the same matrix. 

To characterize the cross-modal attention distributions, statistical descriptors are extracted from the attention responses:
\begin{equation}
A_i^m=
\begin{cases}
A_{i,:}, & m=opt,\\
A_{:,i}, & m=sar,
\end{cases}
\end{equation}
\begin{equation}
s_i^m=
\big[\max(A_i^m),\,\mu(A_i^m),\,\operatorname{Var}(A_i^m)\big].
\end{equation}
where $A_i^m$ is the attention response vector of the $i$-th channel in modality $m$, and $\mu(\cdot)$ and $\operatorname{Var}(\cdot)$ denote its mean and variance. The maximum captures the strongest cross-modal response, the mean provides a scale-dependent baseline (since the opposite-modality channel count varies across scales), and the variance reflects whether attention is concentrated on a few counterparts or spread uniformly. These statistics compactly describe the strength and concentration of cross-modal attention.

The extracted statistics are subsequently mapped to channel scores via
\begin{equation}
    p_i^m=g(s_i^m),
\end{equation}
where $g(\cdot)$ denotes a lightweight scoring function and $p^m=[p_1^m,\ldots,p_{C_m}^m]$ contains the channel scores. The shared and specific indices are obtained by Top-K selection:
\begin{equation}
\mathcal I_m^{b}=\mathrm{TopK}(p^m,k_b),\qquad b\in\{sh,sp\}
\end{equation}
Following our experimental setting, $k_{sh}=0.25C_m$ and $k_{sp}=0.75C_m$. A STE is adopted: the hard indices are used for branch routing in the forward pass, while gradients are approximated through the continuous scores $p^m$ during backpropagation.
\subsection{Dual-Branch Fusion Module}
After obtaining the shared and specific indices, how to exploit the corresponding components becomes critical. Existing methods typically fuse all channels uniformly, which may introduce redundant information and neglect modality-specific characteristics. To address this, the Dual-Branch Fusion Module (DBF) applies differentiated processing to shared and specific features, enabling deep integration of shared information while preserving modality-specific representations.\\
For each modality $m\in\{opt,sar\}$ and branch $b\in\{sh,sp\}$, DBF obtains the corresponding components by
\begin{equation}
\begin{aligned}
F_m^b&=\operatorname{Gather}_{c}(F_m;\mathcal I_m^b),\\
E_m^b&=\operatorname{Gather}_{c}(E_m;\mathcal I_m^b).
\end{aligned}
\end{equation}
This operation produces the shared and specific spatial features together with their embedded counterparts without additional embedding.

\begin{figure*}[!t]
\centering
\includegraphics[width=\textwidth]{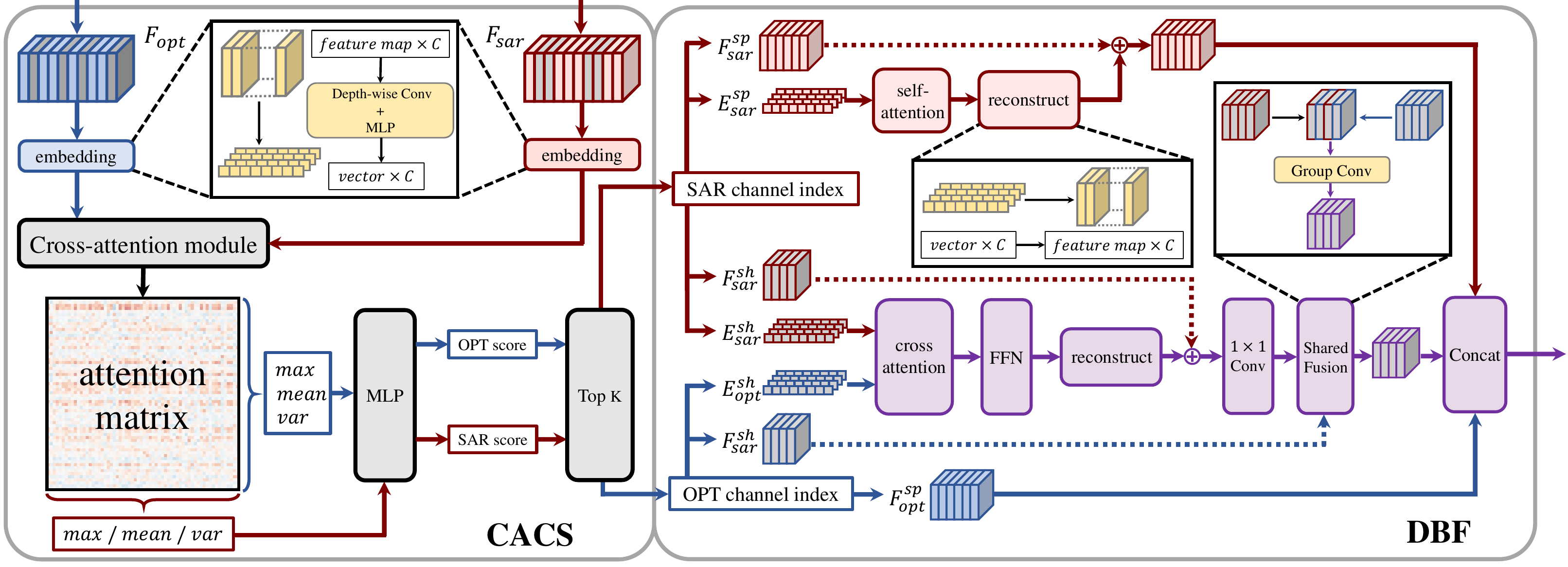} 
\caption{Detailed architectures and data flows of CACS and DBF. Blue, red, and purple denote the optical stream, SAR stream, and fused information or module flow, respectively. CACS constructs the cross-modal attention matrix and selects shared and specific channels through statistical scoring and Top-K selection. DBF performs cross-modal interaction and aligned grouped fusion on shared features, enhances SAR-specific features through self-attention, preserves optical-specific features, and concatenates the three outputs.}
\label{fig:module_details}
\end{figure*}

The proposed DBF consists of two parallel branches: a Shared Feature Fusion Branch and a Specific Feature Preservation Branch, as illustrated in Fig.~\ref{fig:module_details}.

Shared features correspond to channels exhibiting strong cross-modal attention responses, making them suitable for cross-modal interaction. A Transformer-based interaction module therefore operates directly on the shared embeddings selected above. Specifically, the SAR shared embeddings are treated as Query, while the optical shared embeddings serve as Key and Value:
\begin{equation}
E_{sar}^{sh,\ast}
=
\operatorname{Attention}
(
Q_{sar}^{sh},
K_{opt}^{sh},
V_{opt}^{sh}
),
\end{equation}
This lets SAR features selectively aggregate semantically relevant information from the optical modality, enhancing cross-modal representation learning and exploiting complementarity.

The interacted features are then projected back to the spatial domain through a decoder:
\begin{equation}
    \hat F_{sar}^{sh}
=
\Psi(E_{sar}^{sh,\ast}),
\end{equation}
where $\Psi(\cdot)$ denotes the inverse embedding decoder.

To preserve low-level structural details contained in the original shared features, a learnable residual connection is introduced:
\begin{equation}
    \tilde F_{sar}^{sh}
=
\lambda_1
\hat F_{sar}^{sh}
+
F_{sar}^{sh},
\end{equation}
where $\lambda_1$ is a learnable parameter.

Although the shared channels have been identified, potential channel misalignment between modalities may still mix semantically unmatched channels. A $1\times1$ convolution-based channel alignment is therefore applied before fusion:
\begin{equation}
    F_{sar}^{align}
=
\operatorname{Conv}_{1\times1}
\left(
\tilde F_{sar}^{sh}
\right).
\end{equation}
The aligned SAR shared features are then fused with the optical shared features in a pairwise manner:
\begin{equation}
   F^{shared}
=
\Gamma
\left(
F_{opt}^{sh},
F_{sar}^{align}
\right), 
\end{equation}
where $\Gamma(\cdot)$ denotes the shared feature fusion module. Specifically, optical-SAR channel pairs are first constructed. Grouped convolutions are then employed to perform local spatial fusion within each channel pair.

Specific features characterize modality-specific observations that generally lack reliable counterparts, so cross-modal fusion on them may cause information dilution or semantic distortion. Self-attention is therefore applied to the selected SAR-specific embedding before reconstruction:
\begin{equation}
E_{sar}^{sp,\ast}
=
\operatorname{SelfAttn}(E_{sar}^{sp}).
\end{equation}
This operation models dependencies among SAR-specific channels and enhances informative responses without introducing cross-modal interference. The enhanced embedding is then decoded back into the spatial feature domain:
\begin{equation}
    \hat F_{sar}^{sp}
=
\Psi
\big(
E_{sar}^{sp,\ast}
\big).
\end{equation}

A residual connection is further introduced to preserve the original observations:
\begin{equation}
    F_{sar}^{spec}
=
\lambda_2
\hat F_{sar}^{sp}
+
F_{sar}^{sp},
\end{equation}
where $\lambda_2$ denotes a learnable weighting parameter.

Since optical images provide richer texture and structure, their specific features are inherently reliable and are directly retained:
\begin{equation}
    F_{opt}^{spec}
=
F_{opt}^{sp}.
\end{equation}
This design avoids unnecessary cross-modal interference while maximizing the preservation of modality-specific discriminative information.

Finally, the fused shared representation, reconstructed SAR specific features, and preserved optical specific features are concatenated along the channel dimension:
\begin{equation}
F_{out}
=
\operatorname{Concat}
\left(
F^{shared},
F_{sar}^{spec},
F_{opt}^{spec}
\right).
\end{equation}
The resulting representation contains cross-modal shared semantics from deep interaction, modality-specific scattering and structure from SAR, and modality-specific texture and appearance from optical imagery.

Compared with directly concatenating all channels, the proposed strategy retains a single shared representation while preserving all modality-specific information, thereby reducing redundancy and balancing compactness with completeness.


\begin{table*}[t]
\centering
\caption{Comparison with state-of-the-art methods for horizontal object detection on OGSOD 1.0.}
\label{tab:ogsod_sota}
\begin{tabular}{lcccccccccc}
\hline
Method &
Params (M) &
Precision &
Recall &
AP$_{50}$ &
AP$_{75}$ &
AP$_{\text{bridge}}$ &
AP$_{\text{harbor}}$ &
AP$_{\text{storage tank}}$ &
mAP \\
\hline
VSFF~\cite{3_VSFF20243}     & 34.1  & 86.40 & 73.03 & 92.83 & 71.52 & 52.16 & 83.64 & 61.46 & 65.75 \\
CFT~\cite{25_FangCFT}       & 186.4 & 88.36 & 75.69 & \textbf{93.58} & 74.83 & 55.68 & 86.76 & 63.57 & 68.67 \\
CSSA~\cite{19_cao2023CSSA}      & 55.0  & \underline{88.52} & \underline{76.92} & 93.43 & \underline{75.89} & \underline{56.18} & 87.05 & 66.42 & 69.89 \\
ICAFusion~\cite{6_ICAFusion2024} & 99.3  & 87.95 & 76.01 & \underline{93.52} & 75.37 & 56.10 & 87.65 & 65.05 & 69.60 \\
MMIDet~\cite{36_Zeng2024MMIDet}    & 186.4 & 88.13 & 75.57 & 93.44 & 74.40 & 55.75 & 86.67 & 64.41 & 68.94 \\
E2E-OSDet~\cite{24_WangM4SAR} & 56.9  & 88.27 & 76.41 & 93.20 & 75.83 & 53.44 & \underline{88.87} & \underline{67.15} & \underline{69.94} \\
CSIF-Net  & 62.4  & \textbf{89.66} & \textbf{79.32} & 93.33 & \textbf{78.59} & \textbf{58.29} & \textbf{89.74} & \textbf{73.82} & \textbf{73.95} \\
\hline
\end{tabular}
\end{table*}

\section{Experiments} 
\subsection{Datasets and Implementation Details}

We evaluate the proposed method on two public optical-SAR object detection datasets, including OGSOD 1.0 and M4-SAR. These two datasets differ in image resolution, object categories, annotation format, and detection protocol, allowing us to evaluate the proposed fusion strategy under both horizontal and oriented object detection settings.

\textbf{OGSOD 1.0.}
OGSOD 1.0 contains paired SAR and optical images with a spatial resolution of 10 m and an image size of $256 \times 256$. It includes three object categories: bridge, harbor, and storage tank. The dataset is divided into 14,665 training pairs and 3,666 testing pairs. 

\textbf{M4-SAR.}
M4-SAR provides paired optical-SAR images with two spatial resolutions and different SAR polarization modes. The 10 m resolution subset corresponds to VH polarization and contains the first 56,087 image pairs, while the 60 m resolution subset corresponds to VV polarization and contains the remaining 56,087 image pairs. The dataset includes six categories: airport, harbor, bridge, playground, wind turbine, and oil tank. All images are cropped to $512 \times 512$. 

Only the 10 m VH-polarized subset is used, divided into 28,058 training, 11,056 validation, and 16,973 testing pairs. The 10 m and 60 m subsets differ in both resolution and polarization, so mixing them would introduce distribution shifts that obscure the contribution of the proposed fusion module; the 10 m subset thus ensures a consistent distribution and fair comparison.

\textbf{Experimental Settings.} The experiments are organized into horizontal object detection on OGSOD 1.0 and oriented object detection on M4-SAR. For the horizontal setting, ResNet-50, FPN, and TOOD are adopted as the backbone, neck, and detector. Training uses AdamW for 100 epochs with a batch size of 32, a learning rate of $1\times10^{-4}$, a weight decay of $1\times10^{-5}$, and a step schedule with an interval of 10 epochs and a decay factor of 0.6. For the oriented setting, YOLO11n is used with its default backbone and neck. Training uses SGD for 300 epochs with a batch size of 32, an initial learning rate of $1\times10^{-2}$, a weight decay of $5\times10^{-4}$, and a cosine schedule that reduces the learning rate to $1\times10^{-4}$.
\subsection{Horizontal Object Detection}

To evaluate the effectiveness of the proposed CSIF-Net under the horizontal object detection setting, we compare it with several representative multimodal fusion methods on the OGSOD 1.0 dataset, including CFT, CSSA, ICAFusion, MMIDet, E2E-OSDet, and VSFF. For a fair comparison, all methods are implemented under the same detection framework consisting of TOOD with a ResNet-50 backbone and an FPN neck, so that the observed performance differences mainly reflect the effectiveness of different fusion strategies rather than the detector architecture.

VSFF is a pixel-level fusion method that fuses the SAR and optical images before feeding them into a single detection backbone, whereas the remaining methods are feature-level approaches that interact with the two modalities within dual-stream backbones.

\begin{table}[!t]
\centering
\caption{Comparison with state-of-the-art methods for oriented object detection on the M4-SAR 10 m subset.}
\label{tab:m4sar_sota}
\setlength{\tabcolsep}{3.3pt}
\renewcommand{\arraystretch}{1.05}
\begin{tabular}{@{}lcccccc@{}}
\hline
Method &
Params (M) &
Prec. &
Rec. &
AP$_{50}$ &
AP$_{75}$ &
mAP \\
\hline
VSFF~\cite{3_VSFF20243}      & 9.3  & 81.65 & 81.93 & 85.31 & 72.43 & 61.39 \\
CFT ~\cite{25_FangCFT}       & 53.8 & 83.45 & \underline{82.64} & 88.40 & 76.39 & 65.43 \\
CSSA ~\cite{19_cao2023CSSA}     & 13.6 & 82.91 & 82.13 & 87.25 & 75.30 & 63.07 \\
ICAFusion ~\cite{6_ICAFusion2024} & 29.1 & 82.68 & 82.44 & 87.70 & 76.19 & 64.63 \\
MMIDet ~\cite{36_Zeng2024MMIDet}    & 53.8 & 83.16 & 82.53 & 88.65 & 76.62 & 64.54 \\
E2E-OSDet ~\cite{24_WangM4SAR} & 24.6 & \underline{87.09} & \textbf{83.75} & \underline{89.91} & \underline{79.70} & \underline{68.12} \\
CSIF-Net  & 31.4 & \textbf{94.7} & 82.19 & \textbf{91.82} & \textbf{83.67} & \textbf{71.30} \\
\hline
\end{tabular}
\end{table}

Table~\ref{tab:ogsod_sota} reports the quantitative comparison on the OGSOD 1.0 dataset, where the results are analyzed from the perspectives of detection accuracy, localization reliability, cross-scale robustness, and model complexity. In terms of overall detection accuracy, CSIF-Net achieves the best mAP of 73.95\%, surpassing E2E-OSDet (69.94\%), the strongest optical-SAR fusion method, by 4.01 percentage points. Regarding detection reliability, CSIF-Net simultaneously achieves the highest Precision (89.66\%) and Recall (79.32\%), surpassing the best competing results by 1.14 and 2.40 percentage points, respectively, which suggests that the proposed method reduces false-positive detections while recovering more true targets without a precision-recall trade-off. Although CFT attains the highest AP$_{50}$ (93.58\%), its notably lower AP$_{75}$ (74.83\%) reflects weaker localization under stricter IoU, whereas CSIF-Net achieves the highest AP$_{75}$ (78.59\%), confirming more precise object localization. Regarding object scale, CSIF-Net consistently achieves the highest category-wise AP across all three categories. For large-scale objects, CSIF-Net surpasses the best competing results by 2.11 and 0.87 percentage points, respectively. For the compact storage tank category, whose features are subtle and easily overwhelmed by noise, CSIF-Net reaches 73.82\% AP, exceeding the best competing result (67.15\% from E2E-OSDet) by 6.67 percentage points. This demonstrates that the proposed shared-specific channel partitioning is particularly beneficial for small objects, while still improving large-object detection through more discriminative feature interaction. In terms of model complexity, CSIF-Net has only 62.4M parameters, which is substantially fewer than CFT and MMIDet (186.4M) and comparable to CSSA (55.0M) and E2E-OSDet (56.9M). Despite this moderate parameter scale, CSIF-Net achieves the best mAP, indicating a favorable trade-off between model efficiency and detection accuracy.
\subsection{Oriented Object Detection}

To further evaluate the effectiveness and generalization capability of the proposed CSIF-Net under the oriented object detection setting, experiments are conducted on the M4-SAR dataset. All compared methods are implemented using the same YOLO11n detection framework with identical training settings. 

Table~\ref{tab:m4sar_sota} reports the comparison results, where CSIF-Net again achieves the best detection performance in several metrics. In terms of overall detection accuracy, CSIF-Net achieves the best mAP of 71.30\%, surpassing E2E-OSDet (68.12\%), the strongest optical-SAR fusion method, by 3.18 percentage points. Regarding detection reliability, the gain of CSIF-Net is concentrated in precision rather than recall. CSIF-Net achieves the highest Precision of 94.70\%, surpassing E2E-OSDet (87.09\%) by 7.61 percentage points, whereas its Recall (82.19\%) is slightly lower than that of E2E-OSDet (83.75\%). This profile differs from the horizontal setting, where both metrics improved simultaneously, and indicates that CSIF-Net suppresses false-positive detections at a marginal cost of recall, which is particularly valuable for oriented detection where precise bounding-box regression matters more. Correspondingly, CSIF-Net attains the highest AP$_{50}$ (91.82\%) and AP$_{75}$ (83.67\%), improving over E2E-OSDet by 1.91 and 3.97 percentage points, respectively. In terms of model complexity, CSIF-Net uses 31.4M parameters. Notably, CSIF-Net has a parameter count close to that of ICAFusion (29.1M) yet attains a substantially higher mAP (71.30\% vs.\ 64.63\%), indicating that the performance gain stems from the channel-selective fusion strategy rather than from the parameter scale.

\begin{table}[!t]
\centering
\caption{Ablation study on the OGSOD 1.0 dataset.}
\label{tab:ablation}
\setlength{\tabcolsep}{2.1pt}
\renewcommand{\arraystretch}{1.05}
\begin{tabular}{@{}ccccccccc@{}}
\toprule
\multicolumn{4}{c}{Components}
& \multirow{2}{*}{Params (M)}
& \multirow{2}{*}{Recall}
& \multirow{2}{*}{$\mathrm{AP}_{50}$}
& \multirow{2}{*}{$\mathrm{AP}_{75}$}
& \multirow{2}{*}{$\mathrm{mAP}$} \\
\cmidrule(lr){1-4}
FDM & CACS & PB & FB & & & & & \\
\midrule
   & \checkmark & \checkmark & \checkmark
   & 62.4 & 77.2 & 93.4 & 77.9 & 72.99 \\

\checkmark &   & \checkmark & \checkmark
   & 61.3 & 78.1 & 92.9 & 77.2 & 71.68 \\

\checkmark & \checkmark &   & \checkmark
   & 62.1 & 77.3 & 93.5 & 77.9 & 72.46 \\

\checkmark & \checkmark & \checkmark &  
   & 62.1 & 77.9 & 93.1 & 77.0 & 71.21 \\

\checkmark & \checkmark &   &  
   & 59.8 & 75.8 & 93.2 & 74.5 & 68.20 \\

\checkmark &   &   &  
   & 56.9 & 75.1 & 93.1 & 74.2 & 67.09 \\

   &   &   &  
   & 56.9 & 75.4 & 93.1 & 74.3 & 67.13 \\

\midrule
\checkmark & \checkmark & \checkmark & \checkmark
   & 62.4 & 79.3 & 93.3 & 78.6 & 73.95 \\
\bottomrule
\end{tabular}
\end{table}
\subsection{Ablation Study}

To evaluate the contribution of each component, we conduct ablation experiments on the OGSOD 1.0 dataset, as shown in Table~\ref{tab:ablation}. The ablation follows a two-stage strategy. In the first stage, we remove each module individually, i.e., FDM, CACS, PB, and FB, to isolate the contribution of a single component. However, since these modules cooperate with one another, their effects are not simply additive, and the removal of one module may be partially compensated by the remaining ones. We therefore further perform a cumulative combination ablation that progressively removes the modules from the later stages back to the earlier ones, i.e., from the full model down to the baseline, so as to verify how the components jointly contribute to the overall framework. Here, FDM, CACS, PB, and FB denote the Frequency Decomposition Module, Cross-Attention Channel Selection Module, Preservation Branch, and Fusion Branch, respectively.

In the single-module ablation, removing any component degrades $\mathrm{mAP}$. Removing the shared branch causes the largest drop of 2.74 points (from 73.95\% to 71.21\%), followed by removing CACS (2.27 points) and the specific branch (1.49 points), while removing FDM causes the smallest drop of 0.96 points (to 72.99\%). This suggests that the shared-branch cross-modal interaction and channel selection are the primary contributors to detection accuracy, whereas the specific branch and FDM provide smaller but complementary gains by preserving modality-specific information and enriching the frequency representation, respectively. In terms of localization, the complete model attains the highest $\mathrm{AP}_{75}$ of 78.6\%, and removing any single module lowers $\mathrm{AP}_{75}$, indicating that all components collectively contribute to precise localization.

The cumulative combination ablation further confirms the non-additive cooperation among the modules. Removing both branches while keeping FDM and CACS reduces $\mathrm{mAP}$ to 68.20\%, and further removing either CACS or FDM drops it to about 67.1\%, close to the baseline, indicating that these modules contribute little in isolation but become effective through joint cooperation. Overall, while the $\mathrm{AP}_{50}$ values remain close across all variants, the complete model consistently attains the best Recall (79.3\%), $\mathrm{AP}_{75}$ (78.6\%), and $\mathrm{mAP}$ (73.95\%), demonstrating that the four components complement one another and collectively improve detection.

\begin{figure*}[!t]
\centering
\setlength{\featurecellwidth}{0.098\textwidth}
\subfloat[Feature responses before and after cross-modal fusion.\label{fig:feature_before_after}]{%
\begin{minipage}[t]{0.54\textwidth}
\centering
\scriptsize
\begin{minipage}[c]{\featurecellwidth}\centering OPT\end{minipage}
\begin{minipage}[c]{\featurecellwidth}\centering SAR\end{minipage}
\begin{minipage}[c]{\featurecellwidth}\centering OPT Feature\end{minipage}
\begin{minipage}[c]{\featurecellwidth}\centering SAR Feature\end{minipage}
\begin{minipage}[c]{\featurecellwidth}\centering Fused Feature\end{minipage}
\par\vspace{1mm}
\includegraphics[width=\featurecellwidth,height=\featurecellwidth]{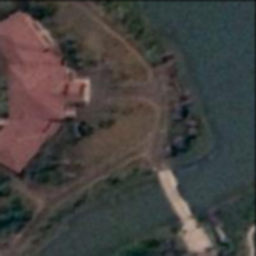}
\includegraphics[width=\featurecellwidth,height=\featurecellwidth]{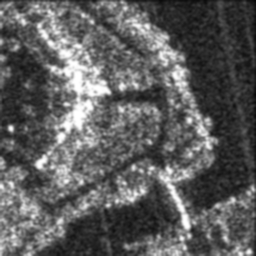}
\includegraphics[width=\featurecellwidth,height=\featurecellwidth]{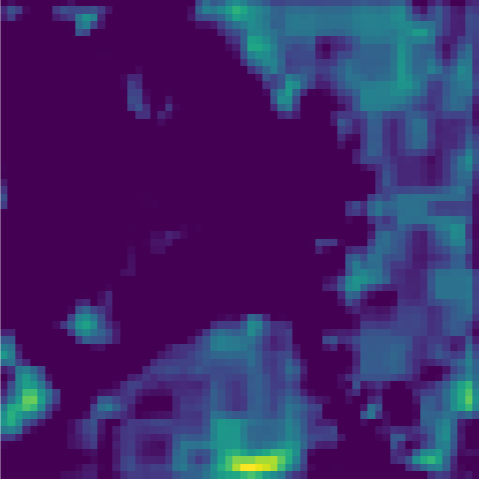}
\includegraphics[width=\featurecellwidth,height=\featurecellwidth]{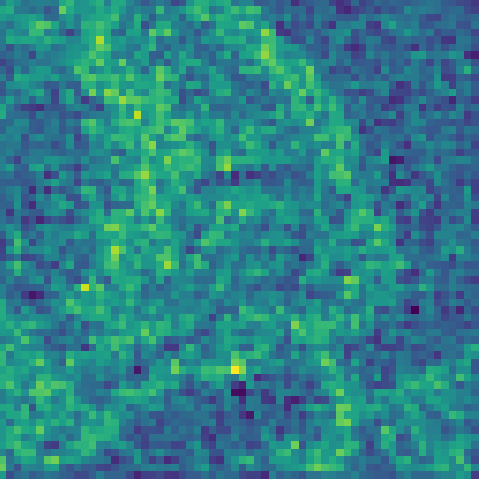}
\includegraphics[width=\featurecellwidth,height=\featurecellwidth]{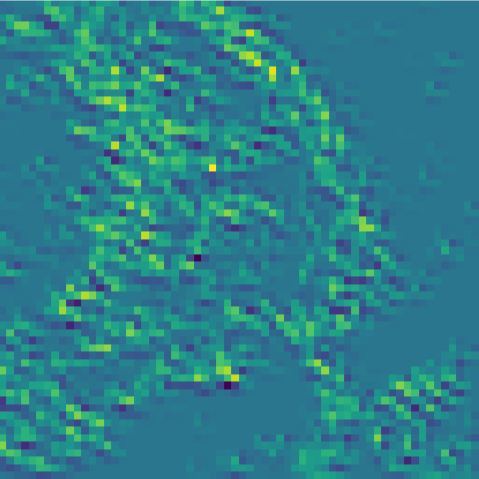}
\par\vspace{0.2mm}
\includegraphics[width=\featurecellwidth,height=\featurecellwidth]{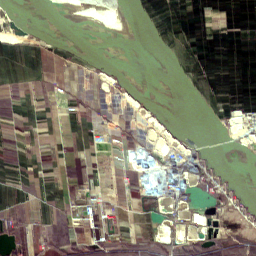}
\includegraphics[width=\featurecellwidth,height=\featurecellwidth]{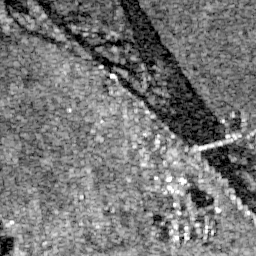}
\includegraphics[width=\featurecellwidth,height=\featurecellwidth]{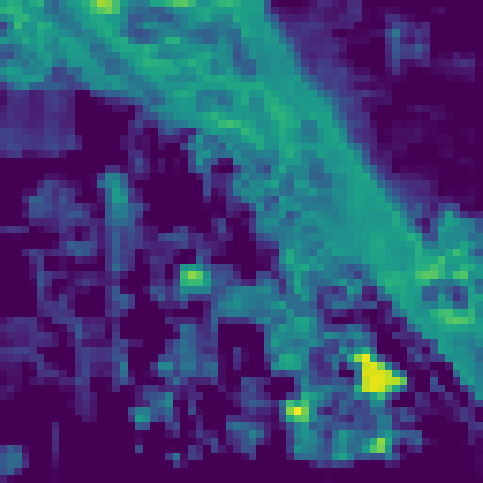}
\includegraphics[width=\featurecellwidth,height=\featurecellwidth]{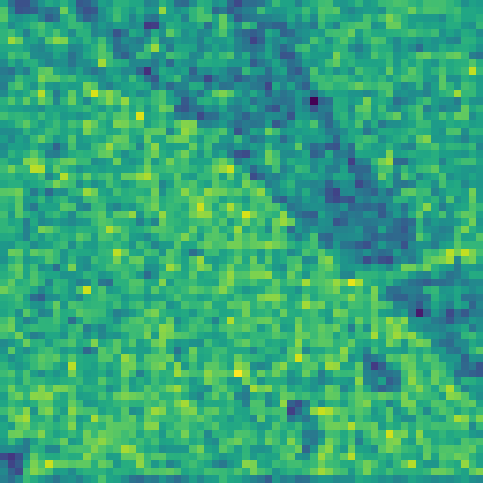}
\includegraphics[width=\featurecellwidth,height=\featurecellwidth]{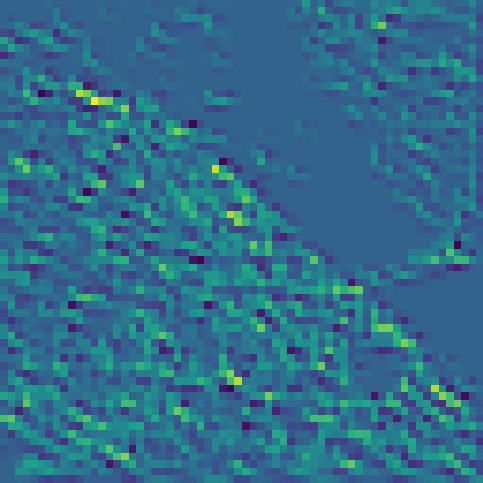}
\par\vspace{0.2mm}
\includegraphics[width=\featurecellwidth,height=\featurecellwidth]{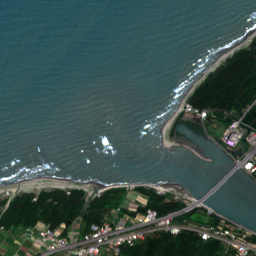}
\includegraphics[width=\featurecellwidth,height=\featurecellwidth]{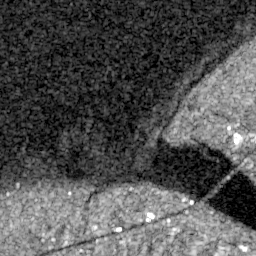}
\includegraphics[width=\featurecellwidth,height=\featurecellwidth]{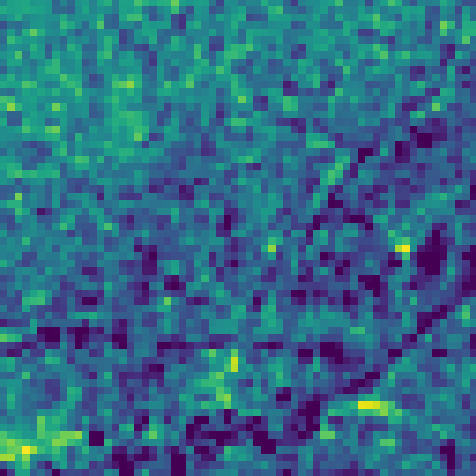}
\includegraphics[width=\featurecellwidth,height=\featurecellwidth]{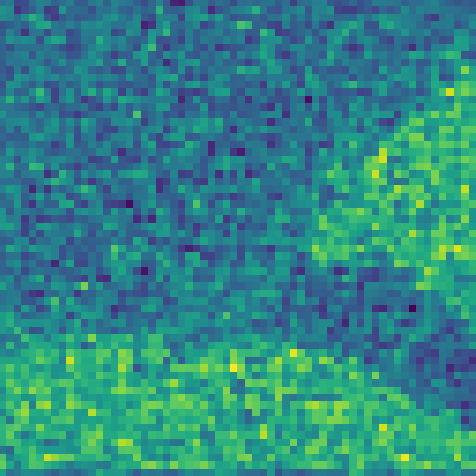}
\includegraphics[width=\featurecellwidth,height=\featurecellwidth]{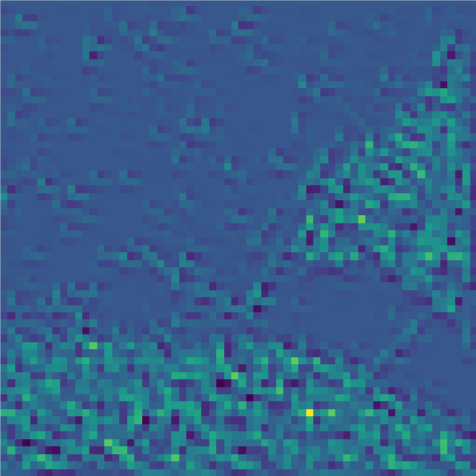}
\end{minipage}}
\hfill
\subfloat[Full-channel versus selected-channel fusion.\label{fig:shared_vs_full}]{%
\begin{minipage}[t]{0.44\textwidth}
\centering
\scriptsize
\begin{minipage}[c]{\featurecellwidth}\centering OPT\end{minipage}
\begin{minipage}[c]{\featurecellwidth}\centering SAR\end{minipage}
\begin{minipage}[c]{\featurecellwidth}\centering Full-channel\end{minipage}
\begin{minipage}[c]{\featurecellwidth}\centering Selected-channel\end{minipage}
\par\vspace{1mm}
\includegraphics[width=\featurecellwidth,height=\featurecellwidth]{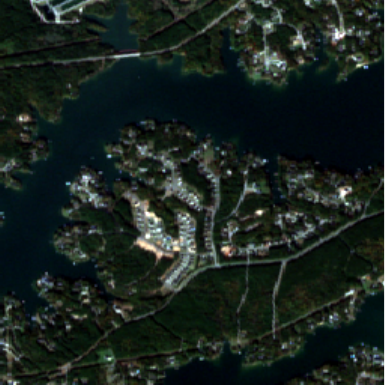}
\includegraphics[width=\featurecellwidth,height=\featurecellwidth]{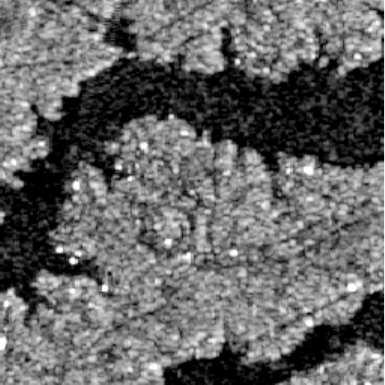}
\includegraphics[width=\featurecellwidth,height=\featurecellwidth]{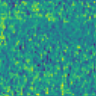}
\includegraphics[width=\featurecellwidth,height=\featurecellwidth]{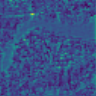}
\par\vspace{0.2mm}
\includegraphics[width=\featurecellwidth,height=\featurecellwidth]{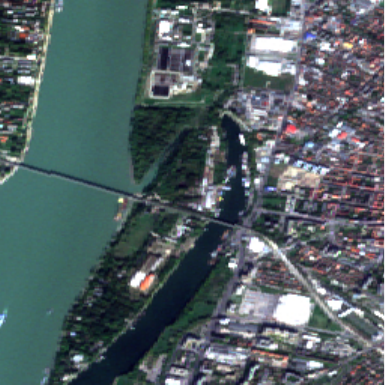}
\includegraphics[width=\featurecellwidth,height=\featurecellwidth]{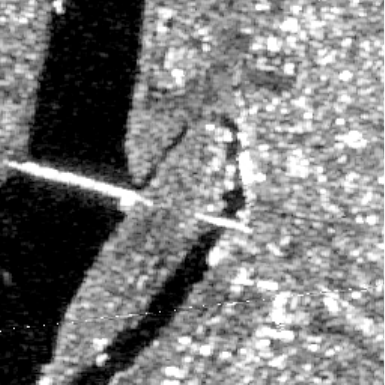}
\includegraphics[width=\featurecellwidth,height=\featurecellwidth]{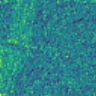}
\includegraphics[width=\featurecellwidth,height=\featurecellwidth]{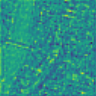}
\par\vspace{0.2mm}
\includegraphics[width=\featurecellwidth,height=\featurecellwidth]{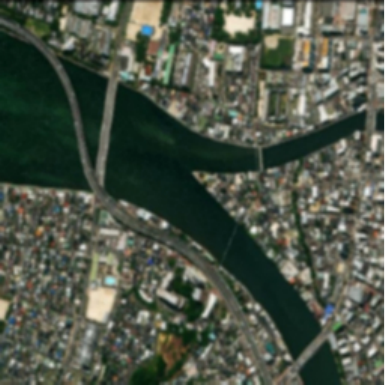}
\includegraphics[width=\featurecellwidth,height=\featurecellwidth]{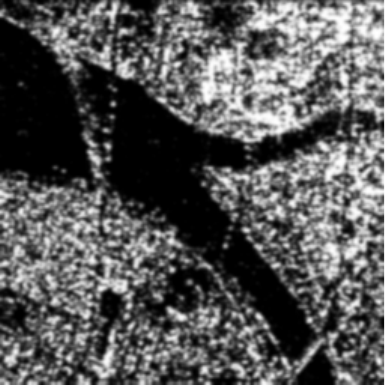}
\includegraphics[width=\featurecellwidth,height=\featurecellwidth]{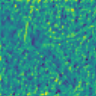}
\includegraphics[width=\featurecellwidth,height=\featurecellwidth]{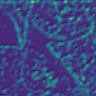}
\end{minipage}}
\caption{Feature-level analysis of CSIF-Net: (a) single-channel responses before and after fusion; (b) full-channel versus selected-channel fusion.}
\label{fig:feature_visualization}
\end{figure*}

\subsection{Model Analysis and Visualization}

To provide an intuitive understanding of CSIF-Net, we visualize intermediate features from two perspectives: single-channel features before and after fusion, and multi-channel features under different fusion strategies.

Fig.~\ref{fig:feature_before_after} shows representative single-channel features before and after fusion. Before fusion, the optical features provide rich appearance and boundary cues, while the SAR features capture structural scattering patterns; the two modalities are thus complementary in content. After channel-selective fusion, the fused features retain the key target information extracted from both modalities while suppressing modality-specific interference: the speckle noise inherent to the SAR modality becomes less salient, so the target-related structures are preserved and the irrelevant background responses are weakened. Moreover, when the optical modality fails to provide reliable cues (for example, when the target color is close to the background or the contours are unclear), the SAR modality can compensate in time by supplying robust structural features, ensuring stable target representation under degraded optical conditions. These observations confirm that CSIF-Net effectively exploits cross-modal complementarity to produce more discriminative features.

Fig.~\ref{fig:shared_vs_full} further compares the feature patterns produced by full-channel fusion and the proposed selected-channel fusion, where the maps are obtained by averaging the ten channels with the largest accumulated gradients. In full-channel fusion, all channels participate in cross-modal interaction, so channels with weak semantic correspondence not only introduce redundant background responses but also bring modality-specific noise and mutual interference from the other modality, diluting the target and contaminating the fused representation. By contrast, selected-channel fusion restricts the cross-modal interaction to the shared channels, preventing the noise-prone and modality-specific channels from disturbing one another, while the remaining channels are preserved in their own branches. As a result, the target responses become more compact and the background is more strongly suppressed. This indicates that the proposed channel selection effectively suppresses both noise and intermodality interference, yielding more discriminative and reliable representations.

\subsection{Qualitative Results}

\begin{figure*}[!t]
\footnotesize
    \centering
    \setlength{\tabcolsep}{1pt}
    \renewcommand{\arraystretch}{1.0}

    \begin{tabular}{@{}*{7}{p{0.139\textwidth}}@{}}
        \centering OPT GT &
        \centering SAR GT &
        \centering CFT &
        \centering CSSA &
        \centering MMIDet &
        \centering E2E-OSDet &
        \centering\arraybackslash CSIF-Net (Ours)
        \\[1mm]

        \includegraphics[width=\linewidth]{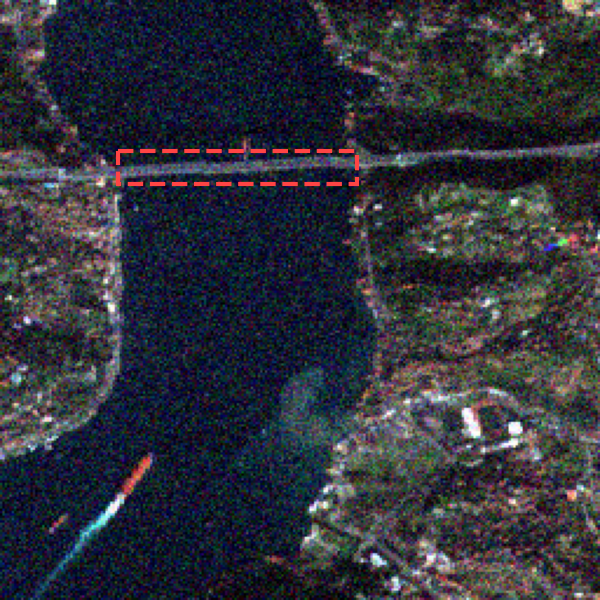} &
        \includegraphics[width=\linewidth]{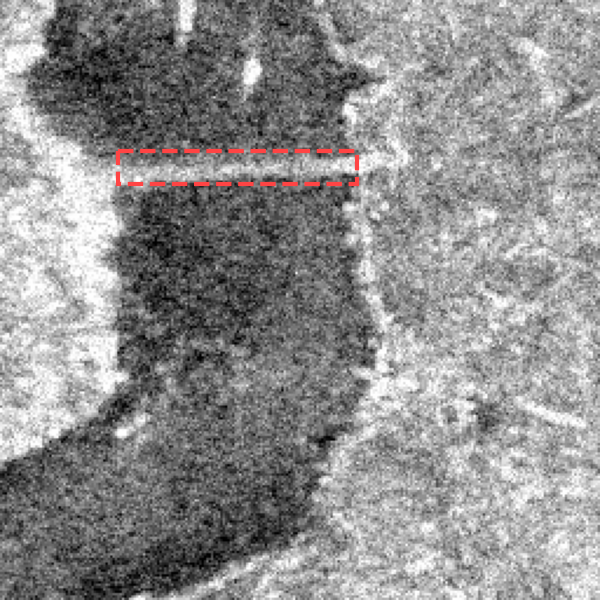} &
        \includegraphics[width=\linewidth]{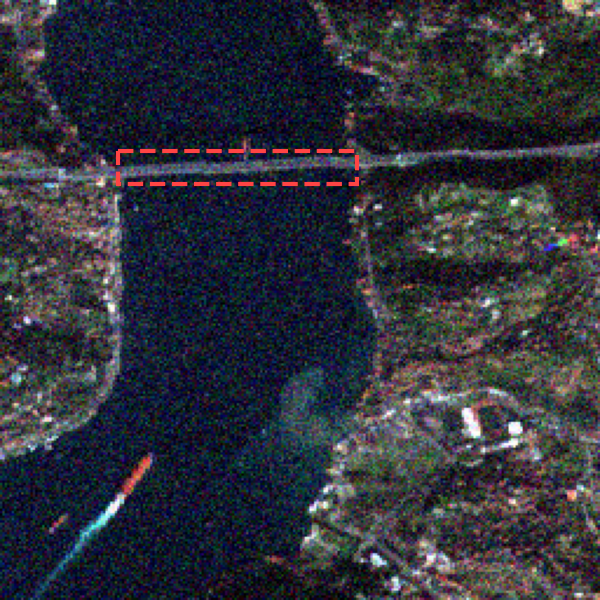} &
        \includegraphics[width=\linewidth]{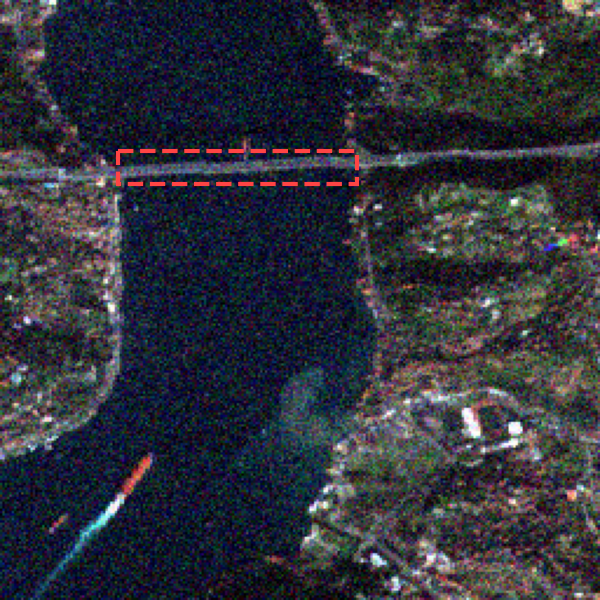} &
        \includegraphics[width=\linewidth]{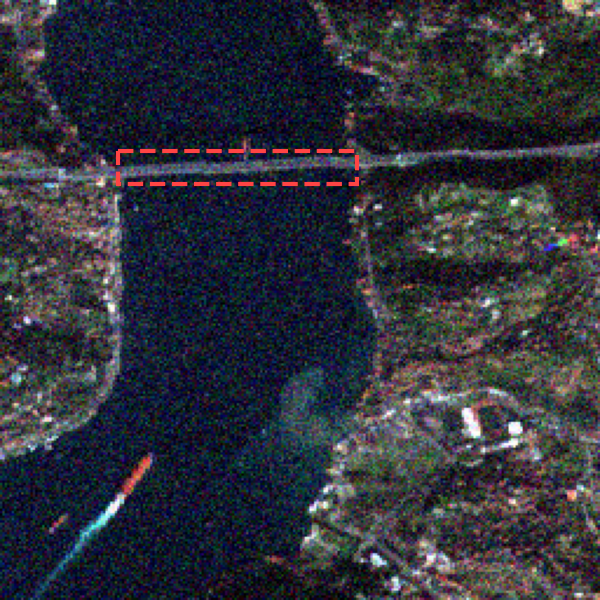} &
        \includegraphics[width=\linewidth]{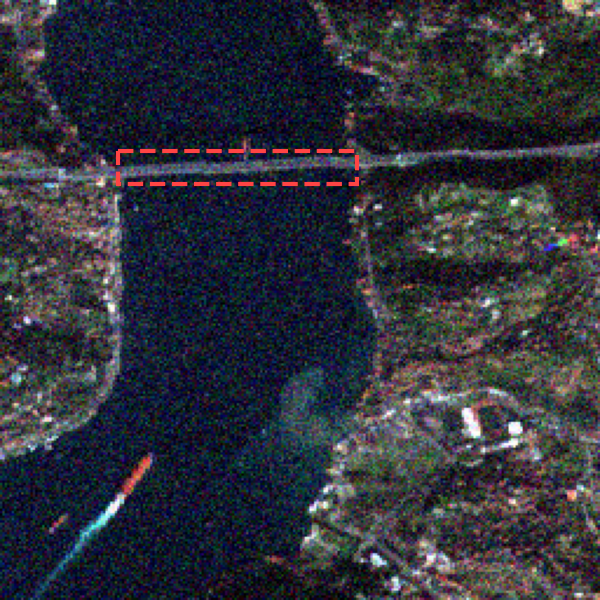} &
        \includegraphics[width=\linewidth]{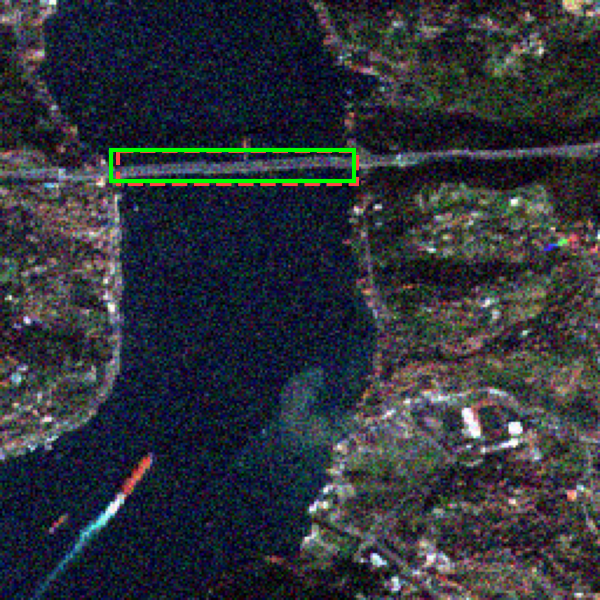}
        \\[1pt]

        \includegraphics[width=\linewidth]{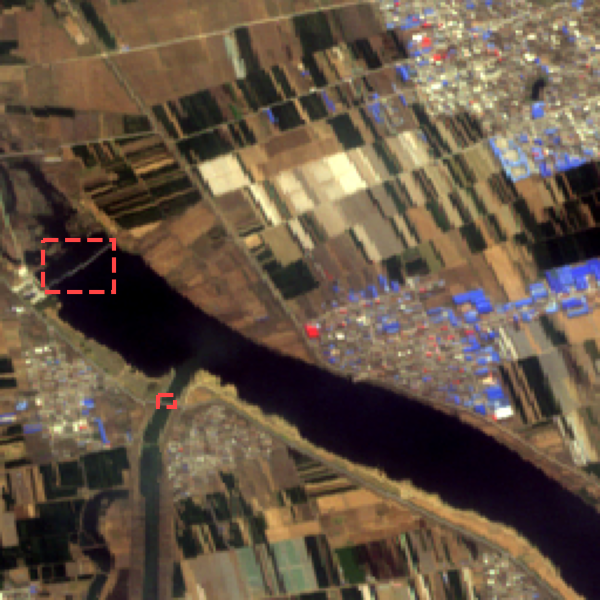} &
        \includegraphics[width=\linewidth]{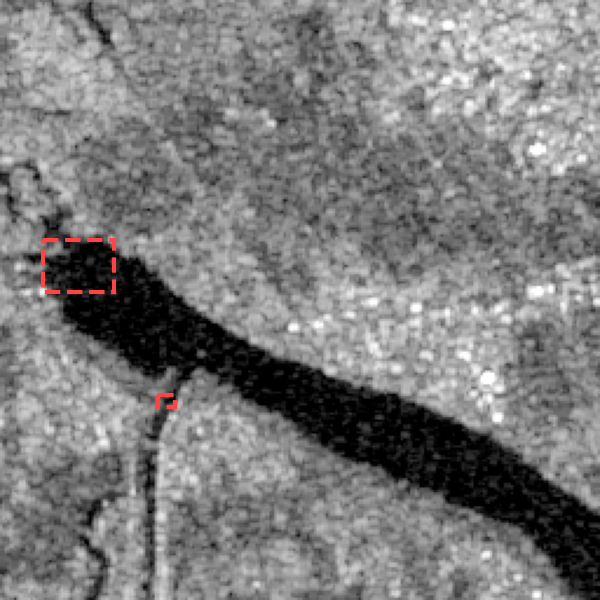} &
        \includegraphics[width=\linewidth]{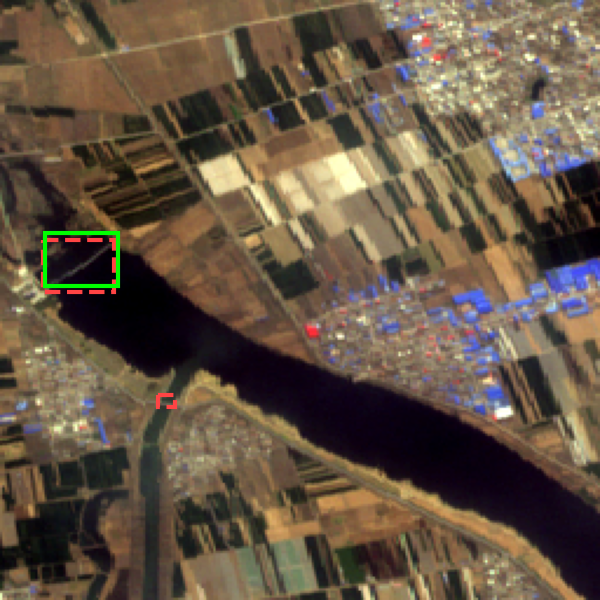} &
        \includegraphics[width=\linewidth]{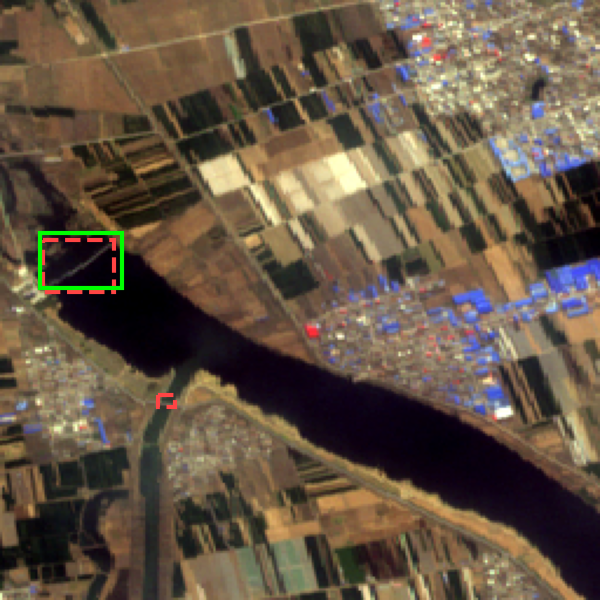} &
        \includegraphics[width=\linewidth]{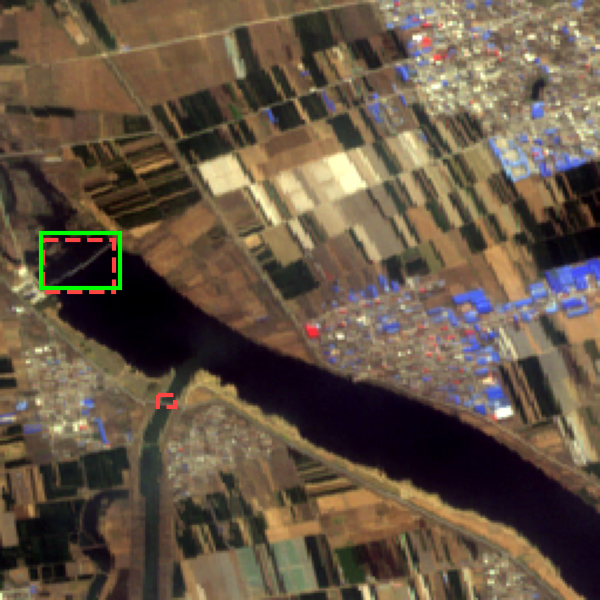} &
        \includegraphics[width=\linewidth]{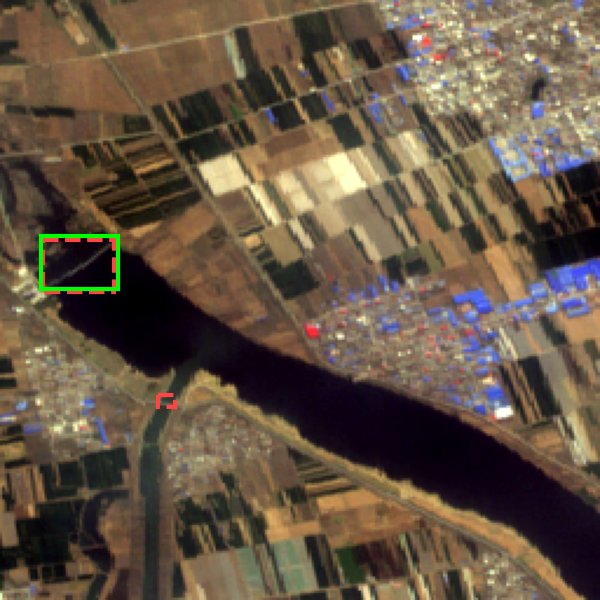} &
        \includegraphics[width=\linewidth]{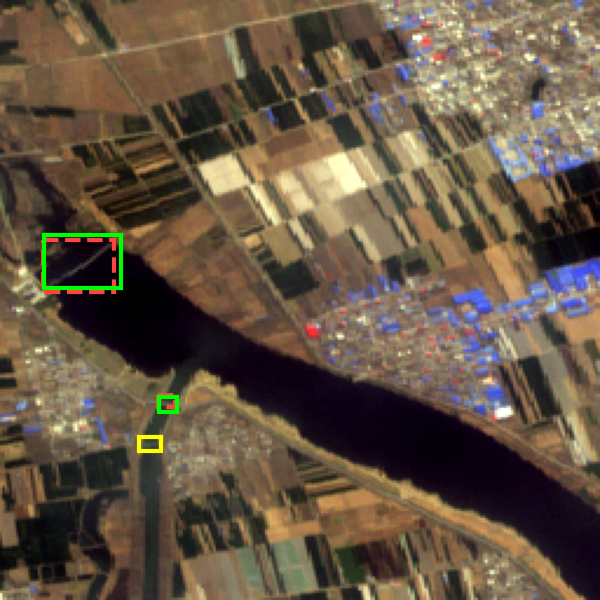}

        \\[2mm]

        \includegraphics[width=\linewidth]{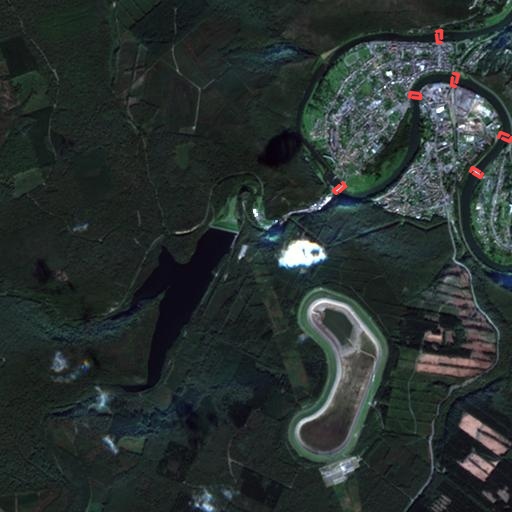} &
        \includegraphics[width=\linewidth]{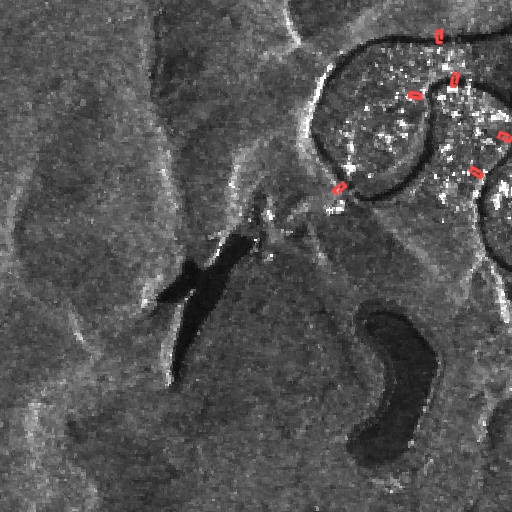} &
        \includegraphics[width=\linewidth]{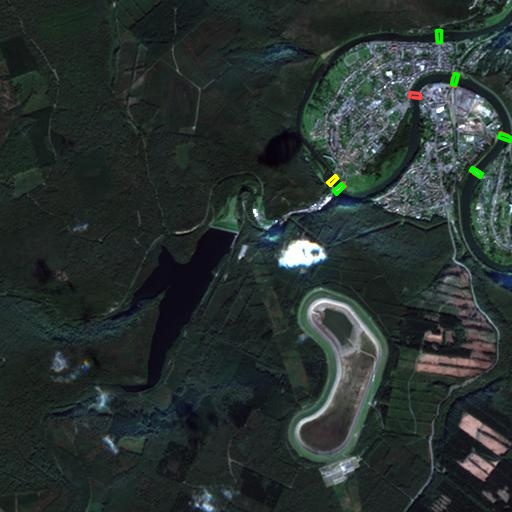} &
        \includegraphics[width=\linewidth]{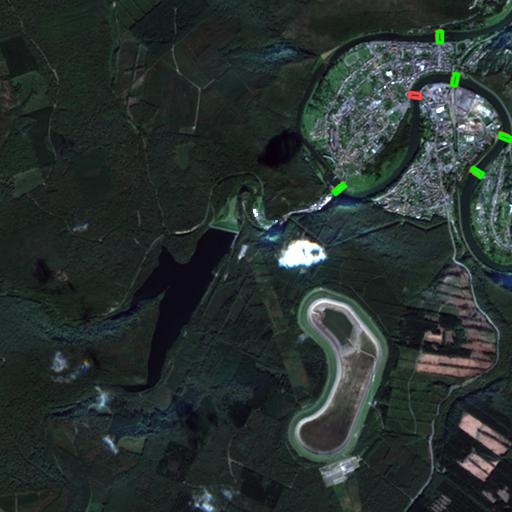} &
        \includegraphics[width=\linewidth]{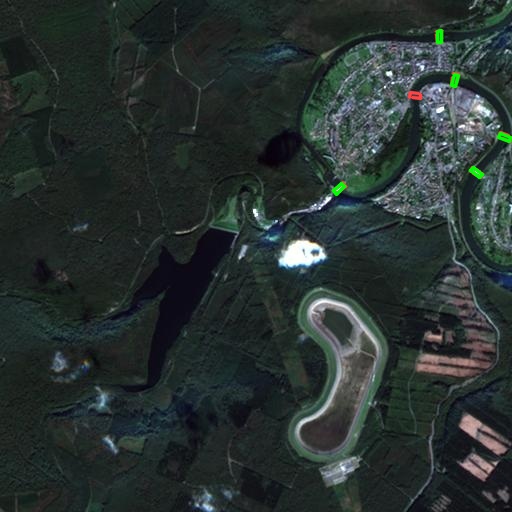} &
        \includegraphics[width=\linewidth]{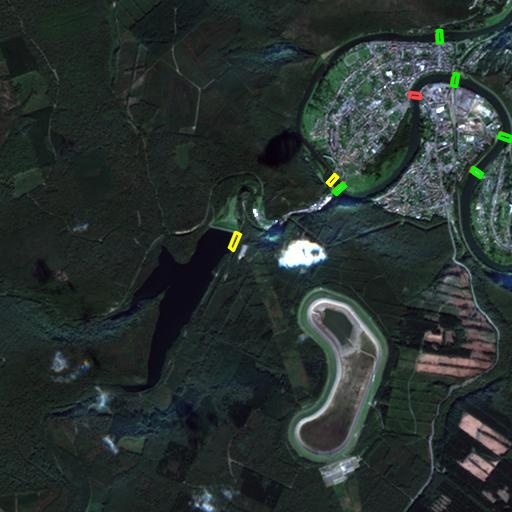} &
        \includegraphics[width=\linewidth]{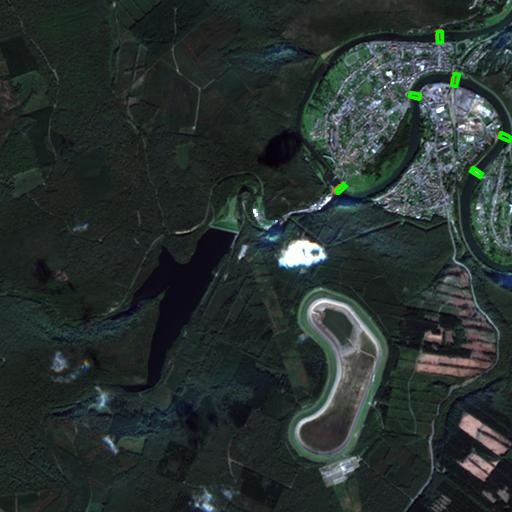}
        \\[1pt]


        \includegraphics[width=\linewidth]{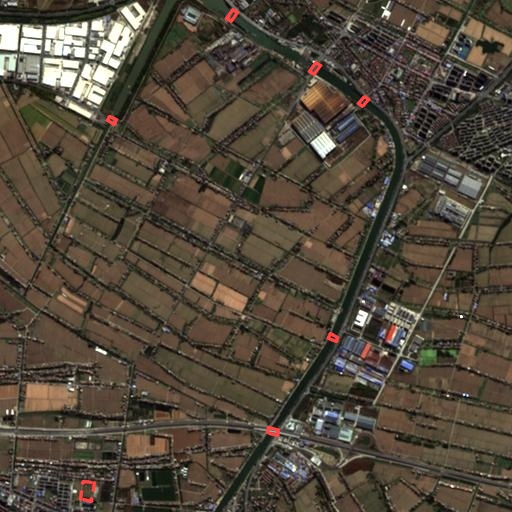} &
        \includegraphics[width=\linewidth]{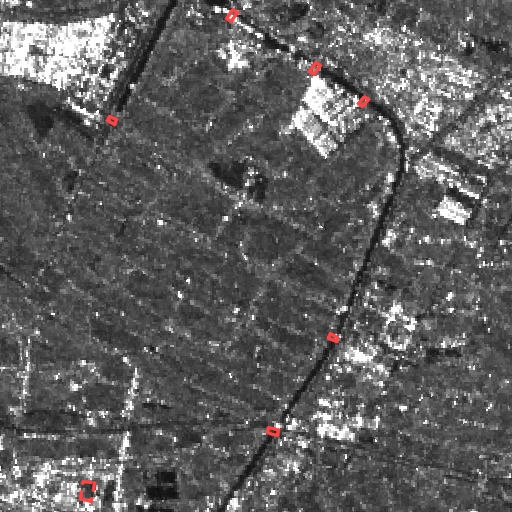} &
        \includegraphics[width=\linewidth]{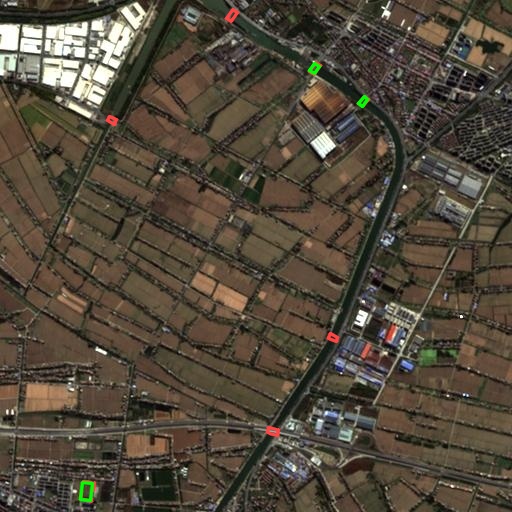} &
        \includegraphics[width=\linewidth]{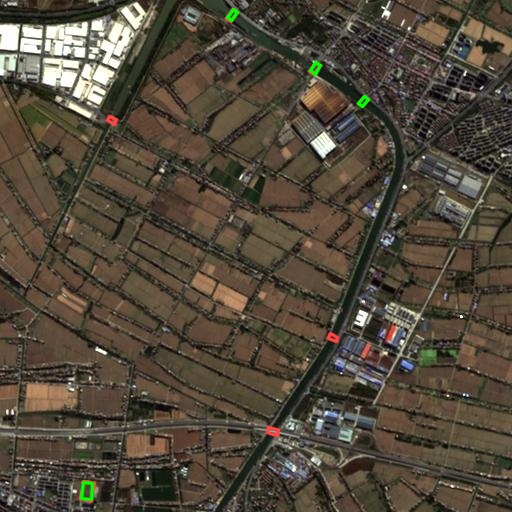} &
        \includegraphics[width=\linewidth]{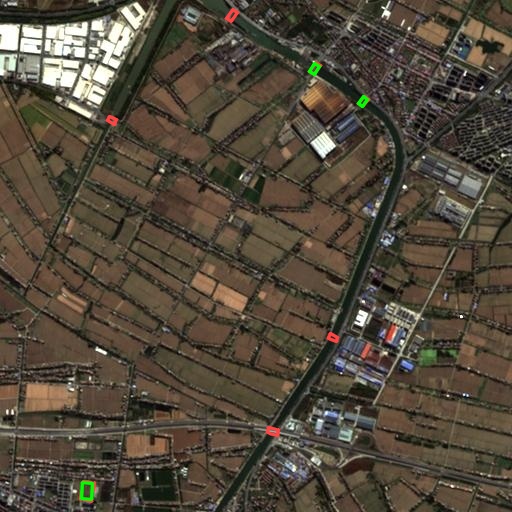} &
        \includegraphics[width=\linewidth]{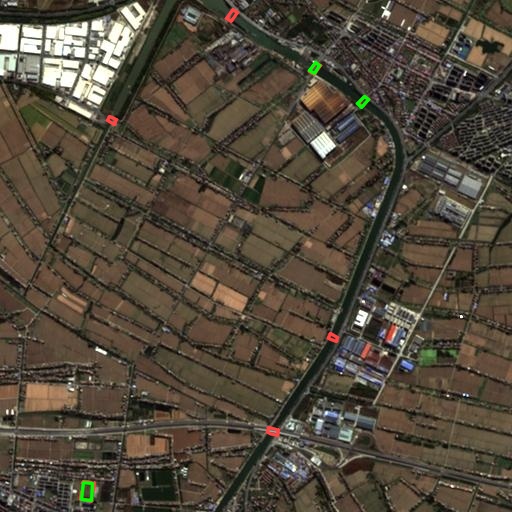} &
        \includegraphics[width=\linewidth]{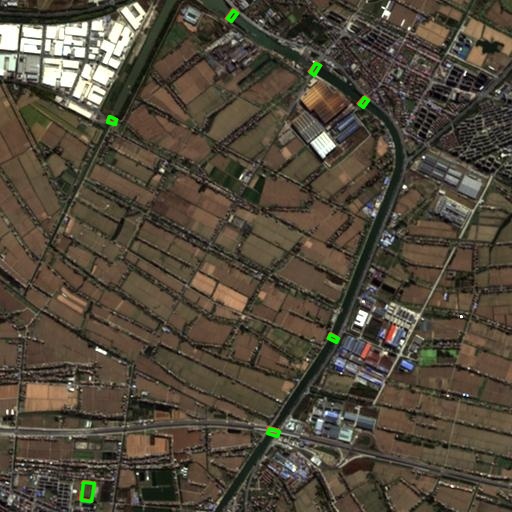}
    \end{tabular}

    \caption{Qualitative comparison of CSIF-Net against existing methods. The top two rows show horizontal bounding-box detection results on representative OGSOD 1.0 scenes, and the bottom two rows show oriented bounding-box detection results on representative M4-SAR scenes. Each row is one paired optical-SAR sample; the first two columns show the optical and SAR ground-truth annotations, and the remaining columns show the optical-view detection results of different methods. Red dashed boxes denote ground truth, yellow solid boxes denote incorrect predictions and green solid boxes denote correct predictions.}
    \label{fig:qualitative}
\end{figure*}

Fig.~\ref{fig:qualitative} presents qualitative comparisons under both the horizontal and oriented object detection settings. The top two rows show horizontal bounding-box results on representative OGSOD 1.0 scenes, and the bottom two rows show oriented bounding-box results on representative M4-SAR scenes. In each row, the paired optical and SAR ground-truth views are shown side by side, followed by the optical-view detection results of the compared methods.

Under the horizontal setting, the main problem is the missed detection of targets with non-salient features or small scales. For such targets, whose responses in the optical modality are weak and easily confused with the background, the compared methods frequently fail to detect them. By exploiting the complementary scattering information from the SAR modality, CSIF-Net recovers these easily missed targets and localizes them more precisely, thereby reducing missed detections.

Under the oriented setting, CSIF-Net mainly benefits from more accurate box orientation and fewer false positives. Methods such as CFT, CSSA, and MMIDet tend to output misaligned boxes or redundant predictions for objects with ambiguous boundaries, whereas CSIF-Net suppresses modality-specific noise and interference through selective channel interaction, yielding clean and well-aligned oriented boxes.

\section{Conclusion} 

This paper presented CSIF-Net, a channel-selective interaction fusion network for optical-SAR object detection. By combining frequency decomposition, cross-modal channel selection, and dual-branch feature fusion, the framework models shared and modality-specific information separately, enabling effective cross-modal interaction while suppressing redundant interference. Experiments on OGSOD 1.0 and M4-SAR show that CSIF-Net outperforms state-of-the-art methods under both horizontal and oriented detection settings, especially for small objects and targets with weak responses in one modality, which is further confirmed by ablation studies and qualitative comparisons.

Despite these gains, CSIF-Net still has two limitations. First, it encodes the raw feature maps directly, which precludes random cropping and other spatial augmentations and thus limits its generalization. Second, the Top-K-based hard selection used to separate the two types of features is non-differentiable and cannot be optimized directly, so it is trained only through a STE. Future work will investigate more adaptive channel interaction mechanisms and extend the framework to more challenging multimodal remote sensing scenarios with additional sensing modalities and larger-scale datasets.

\FloatBarrier
\bibliographystyle{IEEEtran}
\bibliography{IEEEabrv,ref}

\end{document}